\documentclass[runningheads]{llncs}

\usepackage{graphicx}
\usepackage{subcaption}
\usepackage{booktabs} 
\usepackage{multirow}
\usepackage{amsmath, amssymb}
\usepackage{color}
\usepackage{comment}

\newcommand{\argmin}{\operatornamewithlimits{argmin}}
\def\k{\mathbf{k}}
\def\s{\mathbf{s}}

\begin{document}
\pagestyle{headings}
\mainmatter
\def\CVPPPSubNumber{30}  
\def\ECCVSubNumber{\CVPPPSubNumber} 
\title{AutoCount: Unsupervised Segmentation and Counting of Organs in Field Images} 

\titlerunning{AutoCount}
%
\author{Jordan R. Ubbens \and
Tewodros W. Ayalew \and
Steve Shirtliffe \and
Anique Josuttes \and
Curtis Pozniak \and
Ian Stavness}
\authorrunning{J. Ubbens et al.}
\institute{University of Saskatchewan, Saskatoon SK, S7N 5A8, Canada
\email{\{jordan.ubbens,tewodros.ayalew,steve.shirtliffe, \\
acj450,curtis.pozniak,ian.stavness\}@usask.ca}}
\maketitle

\begin{abstract}
Counting plant organs such as heads or tassels from outdoor imagery is a popular benchmark computer vision task in plant phenotyping, which has been previously investigated in the literature using state-of-the-art supervised deep learning techniques. However, the annotation of organs in field images is time-consuming and prone to errors. In this paper, we propose a fully unsupervised technique for counting dense objects such as plant organs. We use a convolutional network-based unsupervised segmentation method followed by two post-hoc optimization steps. The proposed technique is shown to provide competitive counting performance on a range of organ counting tasks in sorghum (\emph{S. bicolor}) and wheat (\emph{T. aestivum}) with no dataset-dependent tuning or modifications.
\keywords{Computer vision, organ counting, unsupervised segmentation.}
\end{abstract}

\section{Introduction}

Object counting is a common real-world use case for deep learning and computer vision. In plant science and agriculture in particular, the counting of plant organs such as heads, tassels, flowers, pods, or fruiting bodies in field environments is relevant, as the number of organs correlates with agronomically important factors such as grain yield. Organ counting is a labor-intensive manual task which is practical to replace with automated methods. There is a substantial body of existing literature describing the counting of organs via supervised learning methods~\cite{Pound2017,Lu2017,Madec2019,Ghosal2019,Sadeghi-tehran2019}. Among the most common supervised techniques are counting by density estimation or local counts regression, object detection, and counting by segmentation. The majority of these supervised methods were originally designed for much more difficult multi-class object detection and object counting problems, such as the challenging COCO benchmark~\cite{lin2014microsoft}. In contrast, plant organs have low intra-class variability, but are dense and overlapping. Supervised methods also require data annotation in the form of either point annotations, bounding boxes, or segmentation masks. Collecting such annotations is difficult and time-consuming, given the typically high density of organs in field images. In addition, poor quality labels can contribute to non-convergence and overfitting as the label noise overpowers the signal in the annotations. Label quality has been shown to be degraded when annotations are provided by domain non-experts, who may be unfamiliar with the plant structures in question~\cite{giuffrida2018citizen}. Principally, these supervised methods need to be retrained on annotated data for each unique dataset, because of the issue of domain shift~\cite{Ubbens2018,Valerio2019}.

\begin{figure}[t]
    \centering
    \includegraphics[width=\textwidth]{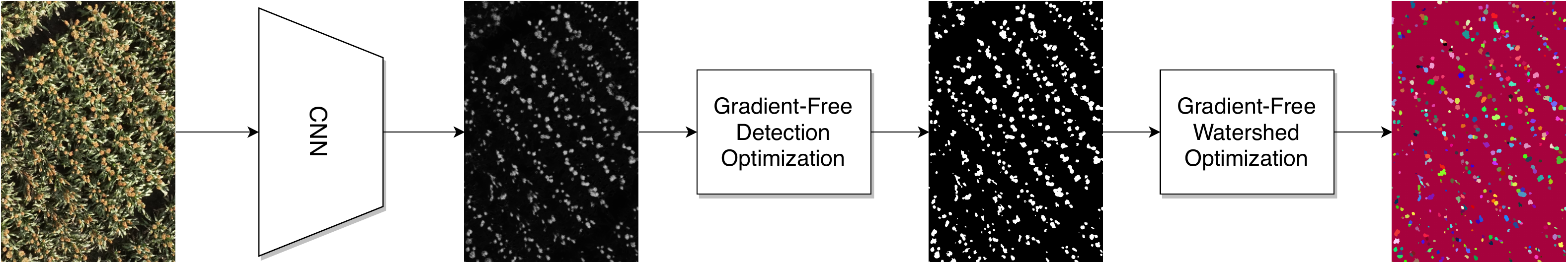}
    \caption{The proposed process from image to instance segmentation.}
    \label{fig:overview}
\end{figure}

In this paper, we propose a novel unsupervised counting technique for counting plant organs in aerial images of fields. We show that a combination of unsupervised segmentation using a CNN, a gradient-free optimization of threshold parameters, and a search over watershed segmentation parameters is able to perform well in these tasks. A complete diagram of the process is shown in Figure~\ref{fig:overview}. We call this proposed object counting system \emph{AutoCount}. We report unsupervised object counting results for two public datasets of sorghum heads (botanical term panicles), as well as a new dataset of wheat heads (botanical term spikes). These results represent new baselines for performing plant organ counting in a fully unsupervised setting, making organ counting easy to apply routinely on a wide variety of aerial image datasets.

\subsection{Related Work}

Organ counting is a common task in image-based plant phenotyping. Early organ counting efforts focused on leaf counting in \emph{A. thaliana} rosettes grown in controlled, indoor conditions such as growth chambers and greenhouses~\cite{Giuffrida2016,Dobrescu2017,ubbens2017deep}. Other work with indoor imaging data has localized wheat spikes and spikelets~\cite{Qiongyan2017,Pound2017}. While performing organ counting with controlled imaging is relevant for many experiments, organ counting is also more relevant in outdoor field-based experiments and is frequently assessed using less-controlled imaging techniques such as with unmanned aerial vehicles (UAVs). Vision tasks in outdoor contexts are substantially different than those in indoor contexts, as factors such as lighting and plant motion cannot be controlled.

Previous work in organ counting in outdoor field environments has explored both image processing as well as deep learning methods. \cite{Guo2018} used a decision tree-based segmentation with an image processing feature set, followed by the classification of regions using Bag-of-Visual-Words features for head detection in sorghum. Deep learning has provided some of the strongest performance for plant organ counting tasks in field imagery. Notably, \cite{Lu2017} used local counts regression on a density map estimated by a CNN. The network showed strong performance for counting the tassels of maize plants from field images. A follow-up work extended the technique to the counting of wheat heads \cite{Xiong2019}. Organ counting has also been explored using object detection methods, such as R-CNN \cite{Hasan2018}, Faster R-CNN \cite{Madec2019}, as well as the state-of-the-art RetinaNet meta-architecture \cite{Ghosal2019}. Others have demonstrated counting by segmentation. \cite{Sadeghi-tehran2019} proposed a superpixel segmentation followed by further segmentation using a fully convolutional network. A similar pipeline has been used for rice panicles \cite{Xiong2017}.

\section{Unsupervised Instance Segmentation}

In contrast to existing organ counting methods, we propose a strictly unsupervised technique which does not require any annotated data. The technique is agnostic to any particular dataset, and does not require any hand-tuning to apply. An open source implementation is available\footnote{\url{https://github.com/p2irc/autocount}}. In the initial step, we make use of an existing unsupervised segmentation technique to train a segmentation network on the entire dataset \cite{Kanezaki2018}. This method utilizes a Simple Linear Iterative Clustering (SLIC) superpixel segmentation to first roughly segment the image. Next, a fully convolutional network consisting of three blocks of only convolution and batch normalization, is used to assign class labels to pixels. This network uses a stride of one on each convolutional layer to maintain the full size of the input image through the network. Pixel-level labels are assigned to superpixels based on the maximum class membership within each region, and the segmentation network is trained on this target using cross-entropy loss. Because of the trivial clustering solution where every region is assigned the same label, intra-axis normalization is used to encourage a larger number of unique class labels. This process is repeated until either a maximum number of iterations or a minimum number of unique class labels has been attained.

Next, the trained segmentation network is used to perform inference on each image in the dataset and the softmax function is applied to the raw activations of the final layer. The activations for the organ class in the output are extracted. The index of this label can either be determined by selecting the label which attains the lowest value of the objective function in Equation~\ref{eq:segmentation-loss}, or it can be indicated by the user. The result is a probability density map for the organ class for each image in the dataset.

In order to obtain a binary segmentation, a gradient-free optimization is performed on the activations using the portfolio-discrete One Plus One optimizer implemented in the Nevergrad library \cite{nevergrad}. We find that this optimization procedure consistently provides more accurate segmentations of object regions than the class predictions from the segmentation CNN. Since the pixel values in this map correspond to probabilities, a log-likelihood loss function is optimized over individual activation maps $\k_i \in \{\k_0 \dots \k_N\}$ as

\begin{equation}
    \argmin_\theta \enspace -{\log\left(\frac{\sum m_\theta (\k_i)}{\sum \k_i}\right)} - \log\left(\frac{\sum m_\theta(\k_i)}{|m_\theta(\k_i)|}\right)
    \label{eq:detection-loss}
\end{equation}

\noindent where $m_\theta$ is a binary masking function which applies a Gaussian filter, a pixel threshold, and then a morphological dilation and erosion to the selected areas. This function returns its input masked by the selected pixels. $|m_\theta|$ indicates the number of selected pixels in that masking, and $\theta = (\tau, \lambda, \phi_{erosion}, \phi_{dilation})$ are the standard deviation of the Gaussian filter, the pixel threshold value, and the diameter of a circular kernel for morphological erosion and dilation, respectively. The first term of the loss function encourages the optimization to include as much of the total probability density as possible, while the second term encourages the average probability of selected pixels to be high.

Finally, objects smaller than $a$ pixels are removed, and watershed segmentation is performed using the distance transform of the binary mask with local maxima no closer than $b$ pixels apart used as watershed markers. These parameters are optimized over binary masks as 

\begin{equation}
    \argmin_{a, b} \enspace \frac{1}{N} \sum_{i=0}^{N} \left[ W(\s_i, \hat{\s}_i) + 
    W(\hat{\s}_i, \mathcal{N}_{\hat{\s}_i}) +  
    W(\Gamma_{\hat{\s}_i}, \mathcal{N}_{\hat{\s}_i}) \right]^2
    \label{eq:segmentation-loss}
\end{equation}

\noindent where $\s$ and $\hat{\s}$ are the empirical distributions of object sizes before and following the procedure, $W$ is the Wasserstein distance, $\mathcal{N}_{\hat{\s}} = \mathcal{N}\left(\mu(\hat{\s}),\,\sigma(\hat{\s})^{2}\right)$ is the normal distribution parameterized by $\hat{\s}$, and $\Gamma_{\hat{\s}} = \Gamma\left(\alpha(\hat{\s}),\,\beta(\hat{\s})\right)$ is the gamma distribution parameterized by $\hat{\s}$. This optimization is performed via an exhaustive search, because of the low dimensionality of the search space. In contrast to Equation~\ref{eq:detection-loss} which is optimized on a per-sample basis, Equation~\ref{eq:segmentation-loss} is optimized over the entire dataset. We find that this mitigates outlier samples which would otherwise find suboptimal solutions. Following optimization, the number of objects is reported as the number of object regions in the watershed segmentation. 

\begin{figure}[t]
    \centering
    \begin{subfigure}[t]{0.3\textwidth}
        \centering
        \includegraphics[width=1.6in]{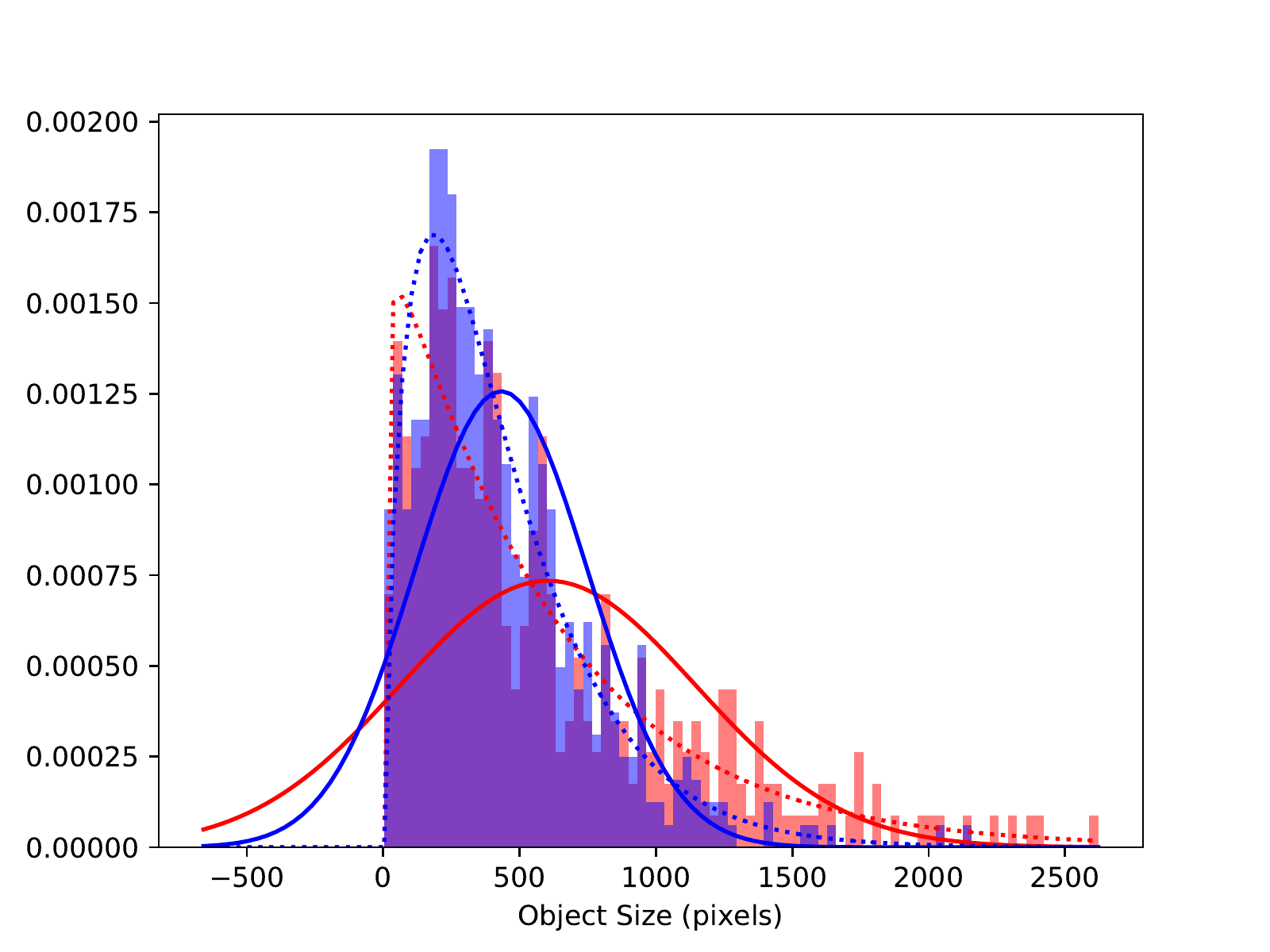}
        \caption{Sorghum dataset 1}
    \end{subfigure}
    ~
    \begin{subfigure}[t]{0.3\textwidth}
        \centering
        \includegraphics[width=1.6in]{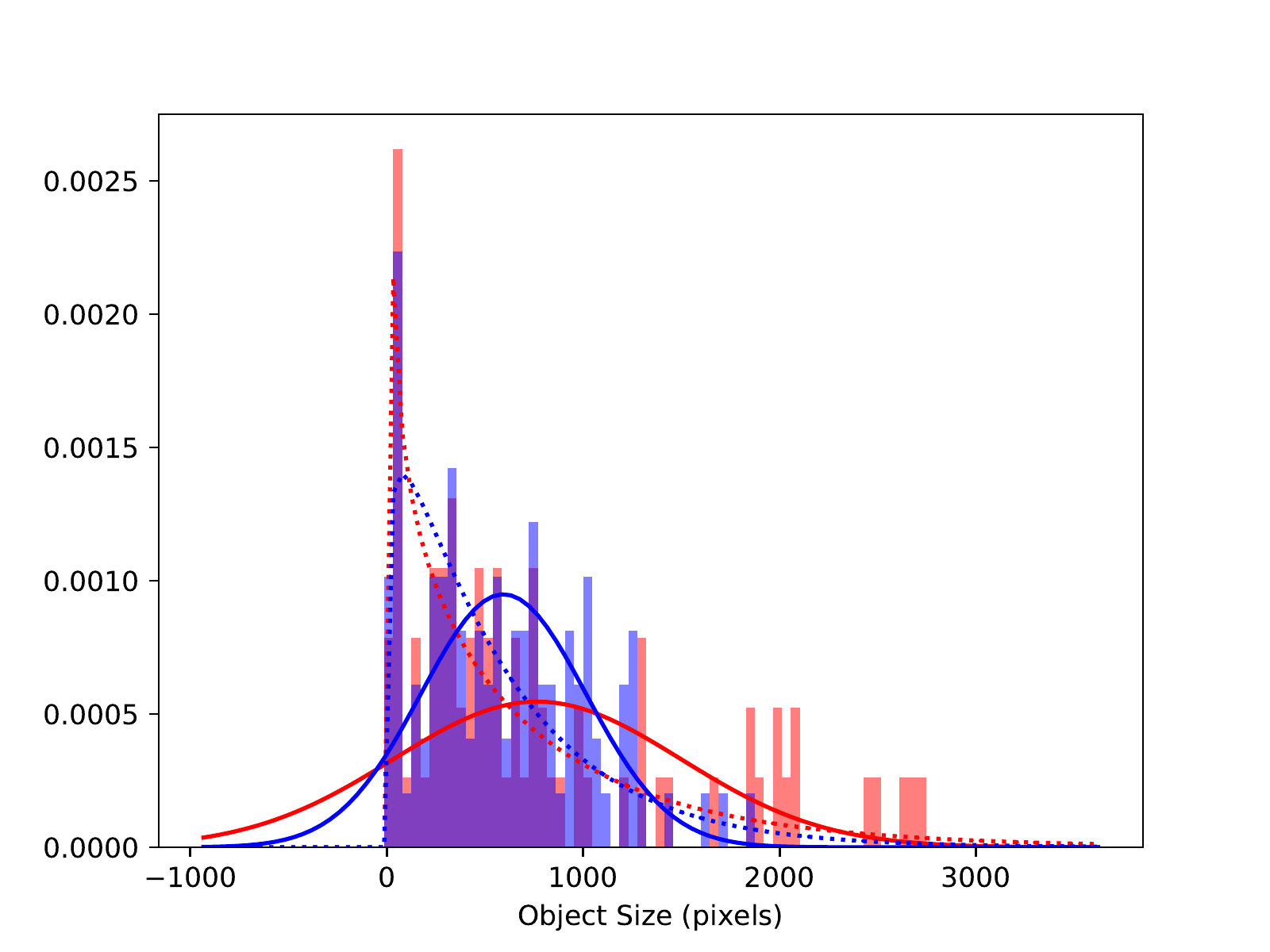}
        \caption{Sorghum dataset 2} 
    \end{subfigure}
    ~
    \begin{subfigure}[t]{0.3\textwidth}
        \centering
        \includegraphics[width=1.6in]{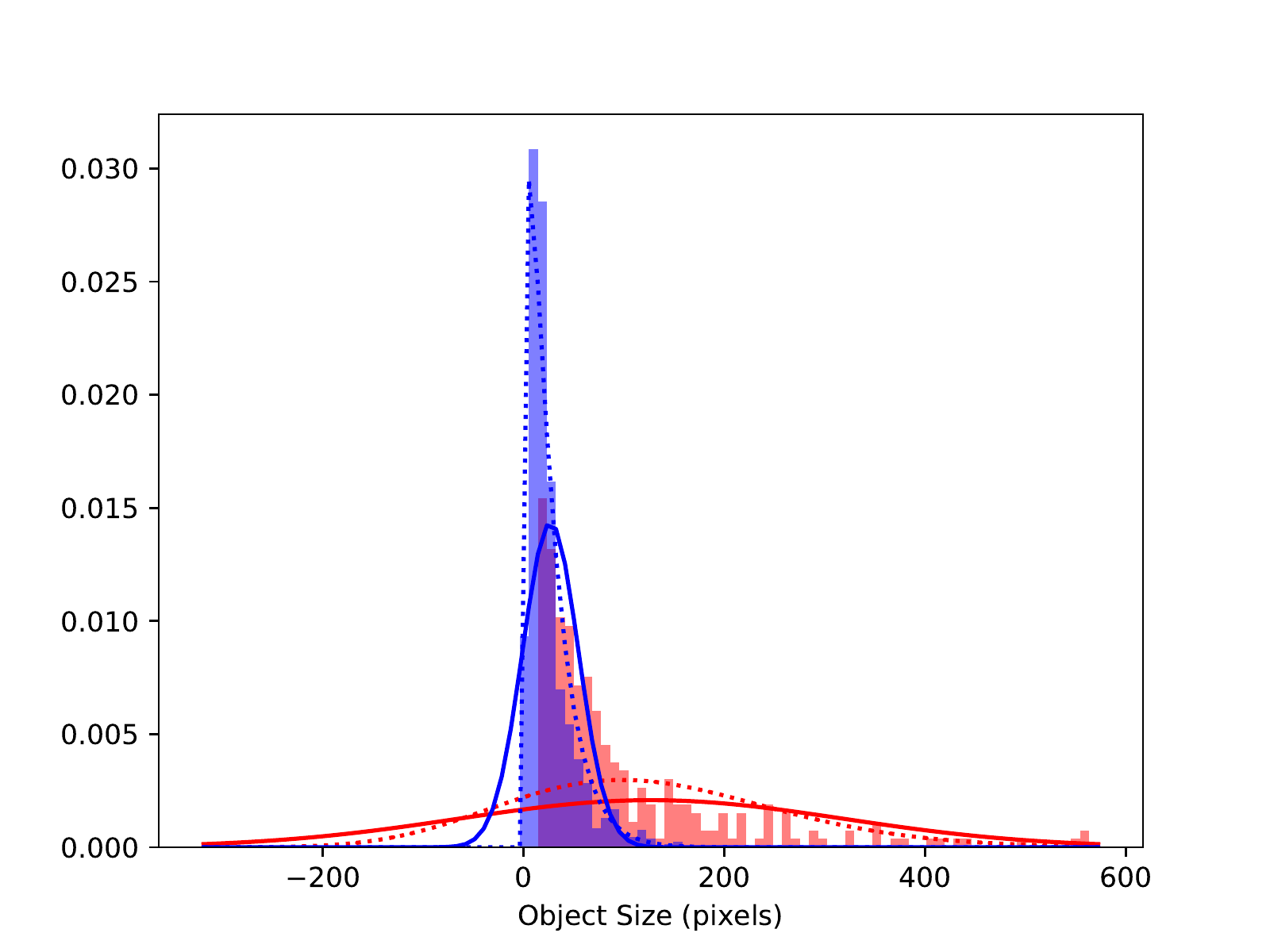}
        \caption{Wheat}
    \end{subfigure}
    
    \caption{Distributions of object sizes for a single example image before (red) and after (blue) the segmentation step. A normal distribution fit to the samples is shown as a solid line and a gamma distribution is shown as a dashed line.}
    \label{fig:distributions}
\end{figure}

Intuitively, for objects such as plant organs from top-down aerial images, we assume that the object sizes should be roughly normally distributed. The tension between the first term, which acts as a prior, and the second two terms, which shape the distribution by affecting its normality and skewness, finds a balance between over-segmentation and under-segmentation of object regions. Note that, if $\s$ is already normally distributed, then the optimal solution to Equation~\ref{eq:segmentation-loss} is to leave the distribution unchanged. Figure~\ref{fig:distributions} shows the distributions of object sizes for the sorghum and wheat experiments before and after this step.

Although the training of the CNN in the initial step includes several tuneable hyperparameters such as the learning rate and the number of convolutional blocks, we use the default settings provided by the authors' implementation for all experiments \cite{Kanezaki2018}. We adjust the number of output channels to 32 and the final number of labels (used as a stopping criteria) to eight, based on the low level of object diversity present in field datasets.

\section{Results}

\begin{figure}[h]
    \centering
    \begin{subfigure}[t]{0.66\textwidth}
        \centering
        \includegraphics[height=1.5in]{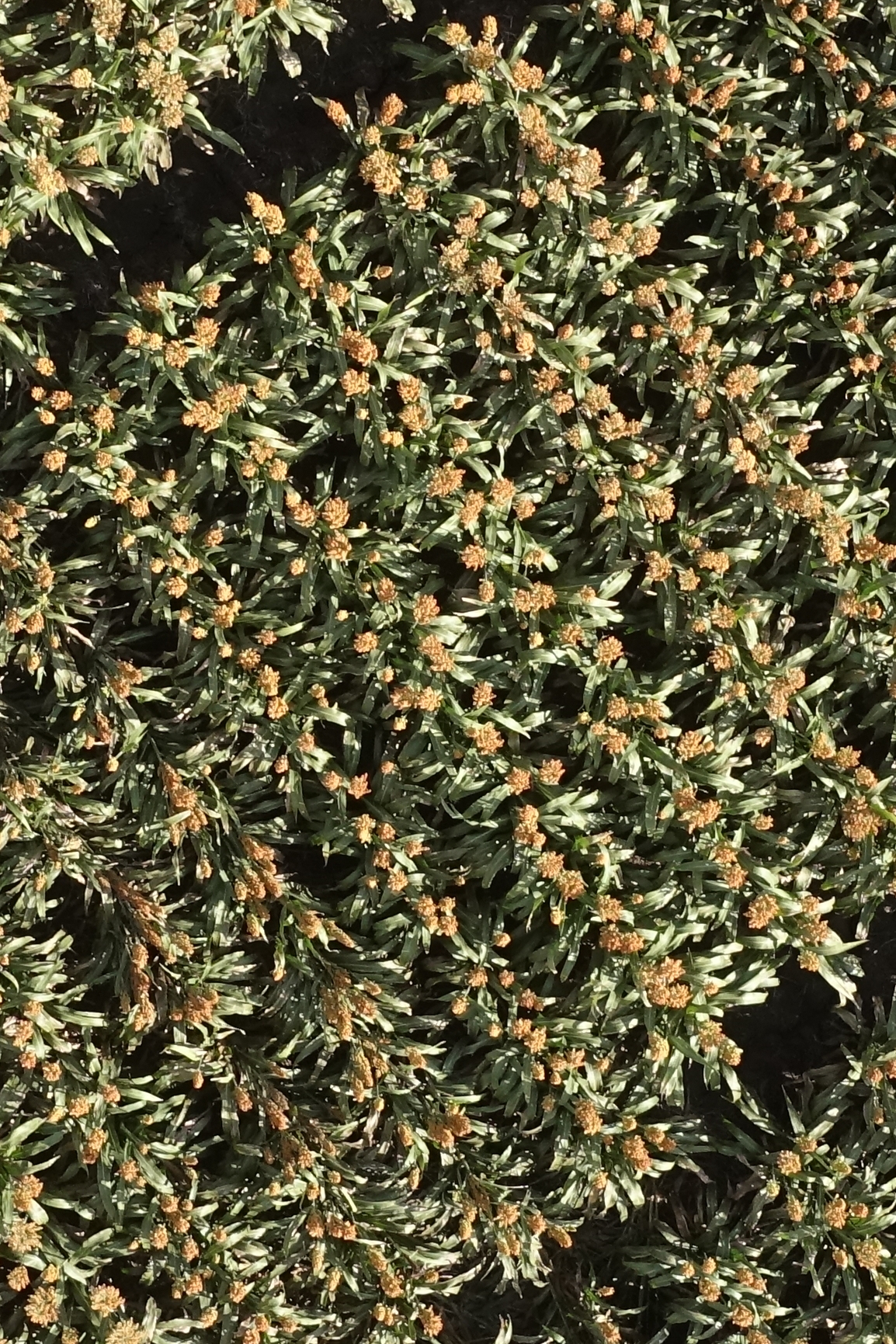}
        \includegraphics[height=1.5in]{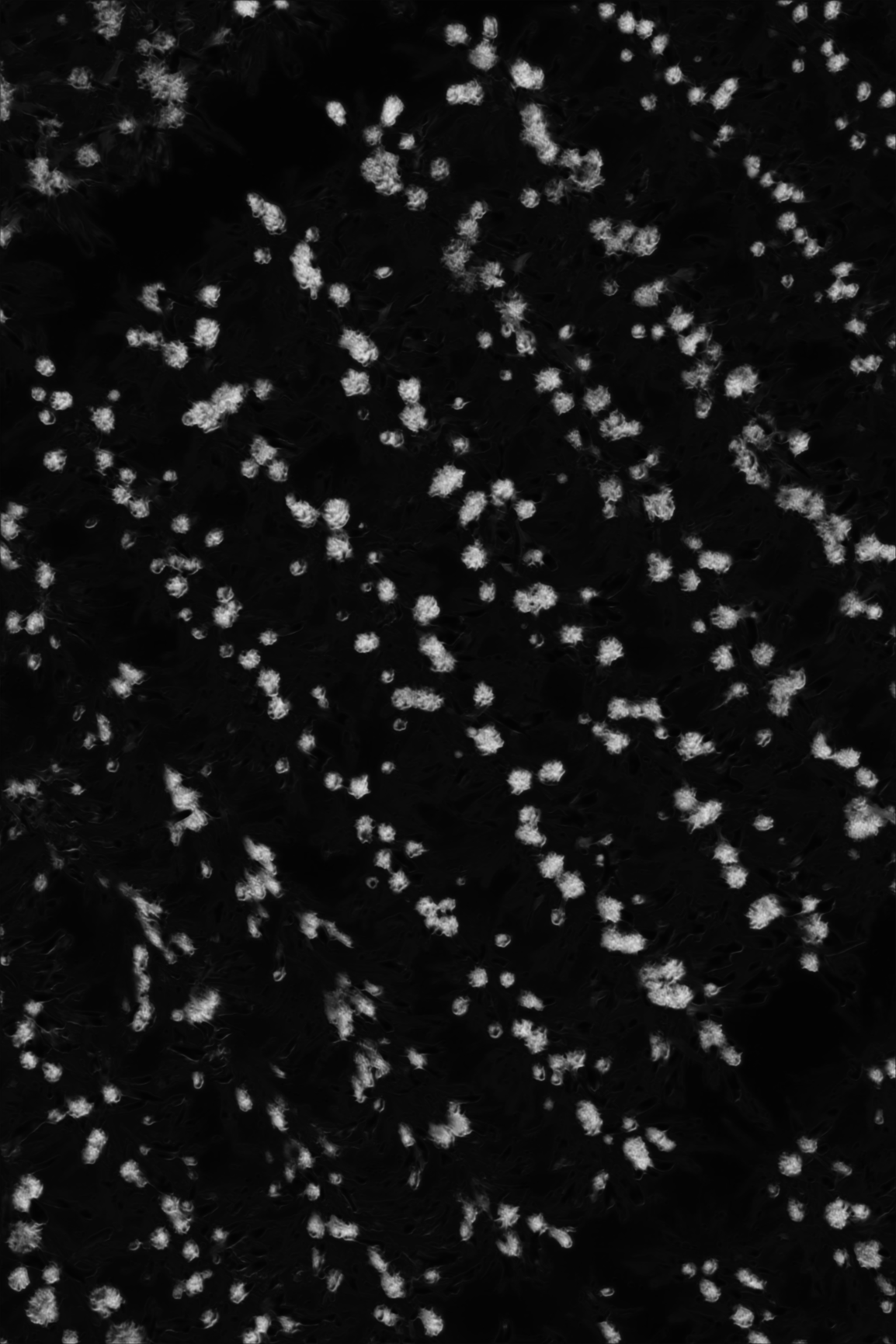}
        \includegraphics[height=1.5in]{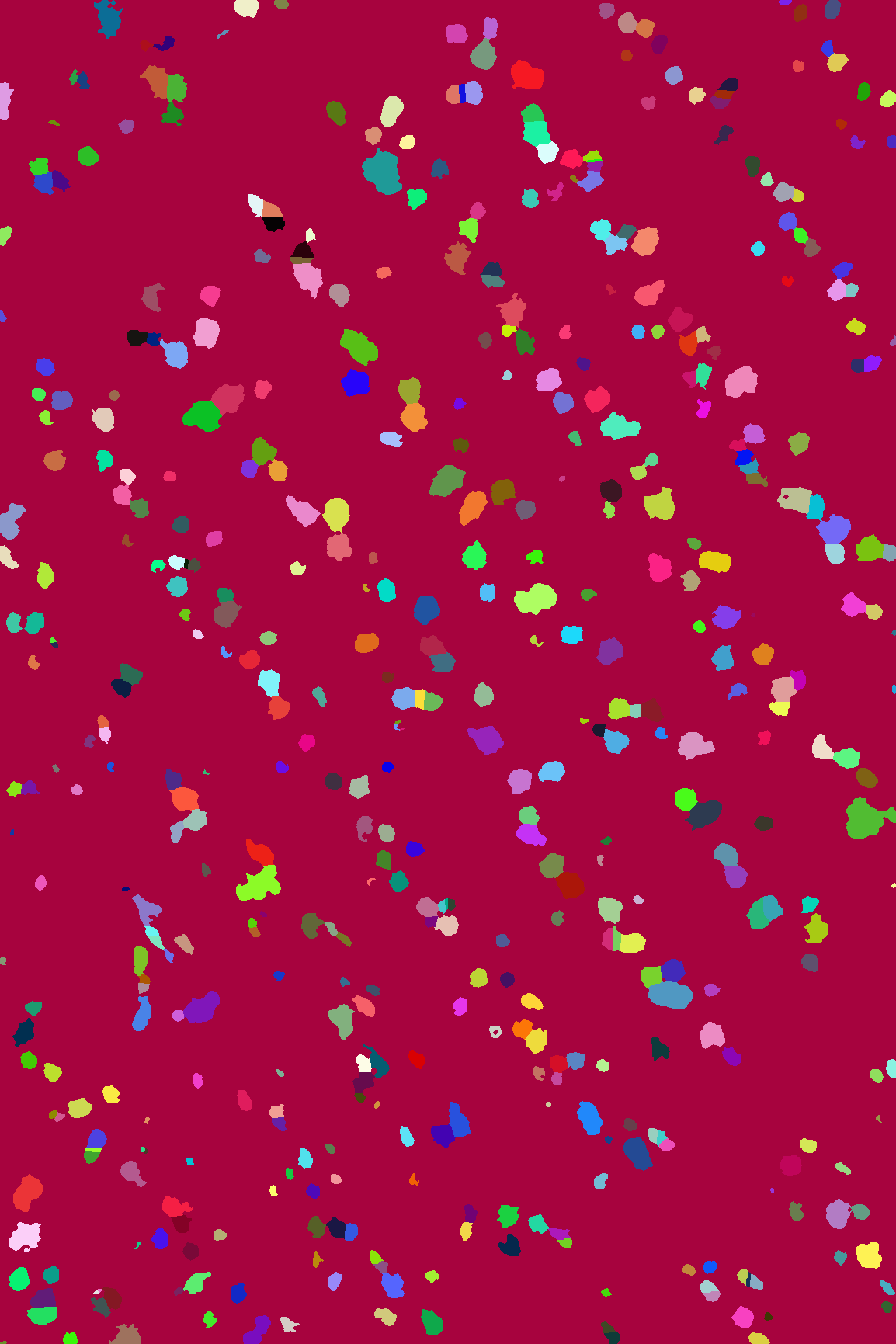}
    \end{subfigure}%
    ~ 
    \begin{subfigure}[t]{0.32\textwidth}
        \centering
        \includegraphics[angle=90,height=1.5in]{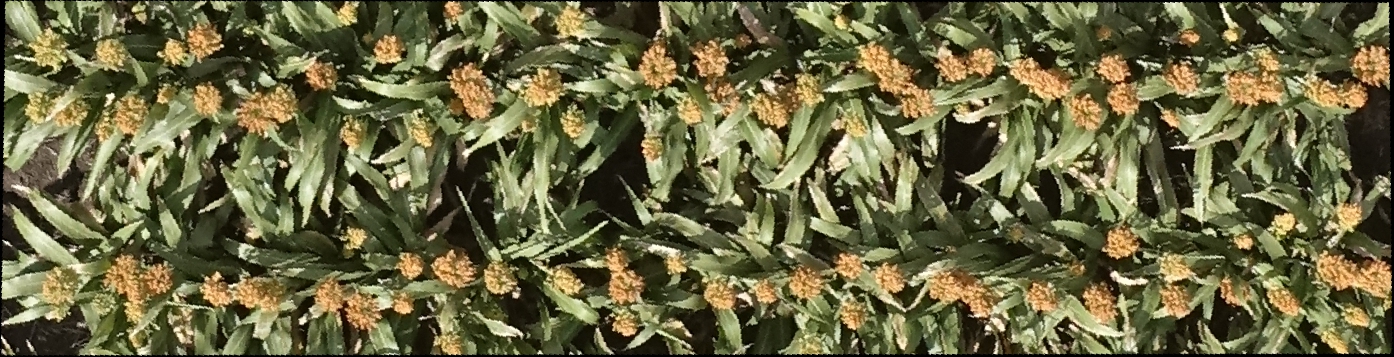}
        \includegraphics[angle=90,height=1.5in]{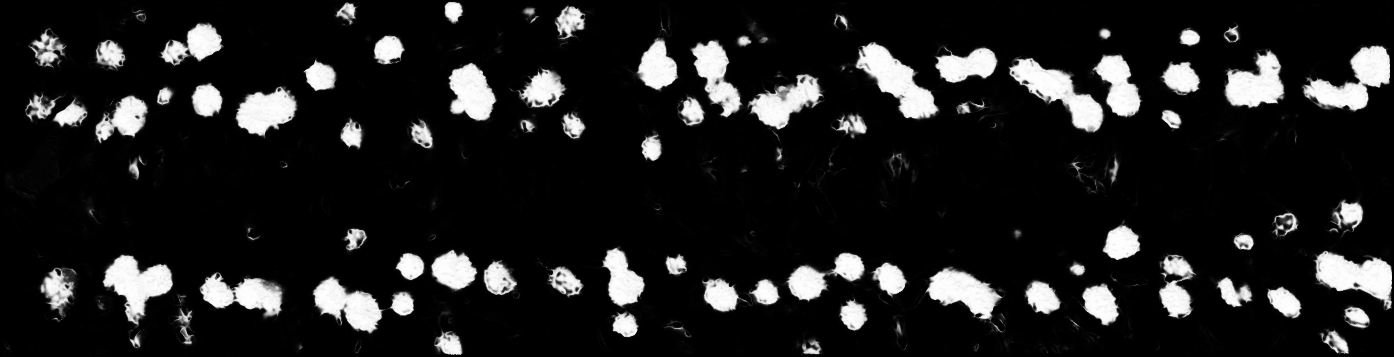}
        \includegraphics[angle=90,height=1.5in]{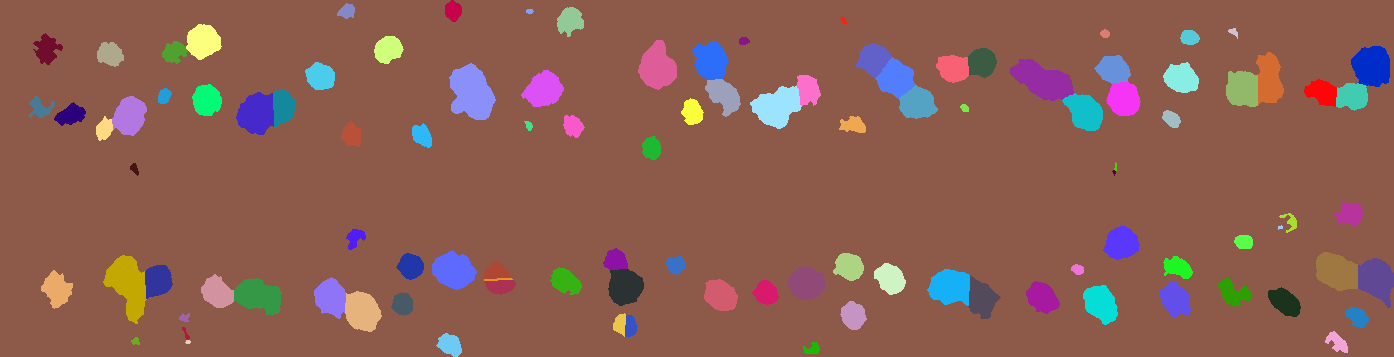}
    \end{subfigure}
    \caption{Example images and intermediate outputs from the sorghum datasets. Left to right: original image, softmax activations, watershed output.}
    \label{fig:examples-sorghum}
\end{figure}

We evaluate the counting performance of the proposed method using the Mean Absolute Error (MAE), the Root Mean Square Error (RMSE), and the coefficient of determination between the annotated and predicted counts ($\text{R}^2$). Table~\ref{tab:results} shows a comparison against two previously reported results from the literature on two publicly available datasets of sorghum heads \cite{Guo2018}. The resolution for these datasets is 1154 by 1731 pixels for dataset 1 and 1394 by 357 pixels for dataset 2. Although counting performance fell short of the previous standard set by the powerful RetinaNet object detection meta-architecture applied in \cite{Ghosal2019}, AutoCount shows that a high degree of performance can be attained without using labels. Examples from the public sorghum datasets are shown in Figure~\ref{fig:examples-sorghum}.

\begin{table}[h]
\caption{Counting results for the sorghum and wheat head datasets.}
\label{tab:results}
\begin{center}
\begin{tabular}{llccc}
& & MAE & RMSE & $\text{R}^2$ \\
\toprule
\multirow{3}{*}{Sorghum 1 \enspace}
&Segmentation + Classification \cite{Guo2018} \enspace & - & - & 0.84 \\
&RetinaNet \cite{Ghosal2019} & - & - & 0.82 \\
&AutoCount (unsupervised) & 29.92 & 36.39 & 0.79 \\
\midrule
\multirow{3}{*}{Sorghum 2}
&Segmentation + Classification \cite{Guo2018} & - & - & 0.56 \\
&RetinaNet \cite{Ghosal2019} & - & - & 0.76 \\
&AutoCount (unsupervised) & 4.95 & 6.22 & 0.48 \\
\midrule
Wheat
&AutoCount (unsupervised) & 297.22 & 344.89 & 0.17 \\
\bottomrule
\end{tabular}
\end{center}
\end{table}

\begin{figure}[h]
    \centering
    \begin{subfigure}[t]{\textwidth}
        \centering
        \includegraphics[width=0.32\textwidth]{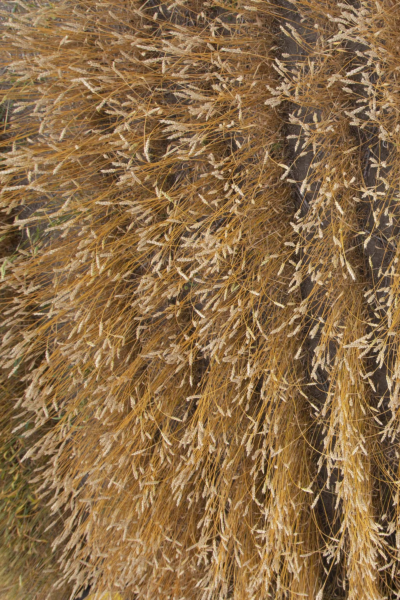}
        \includegraphics[width=0.32\textwidth]{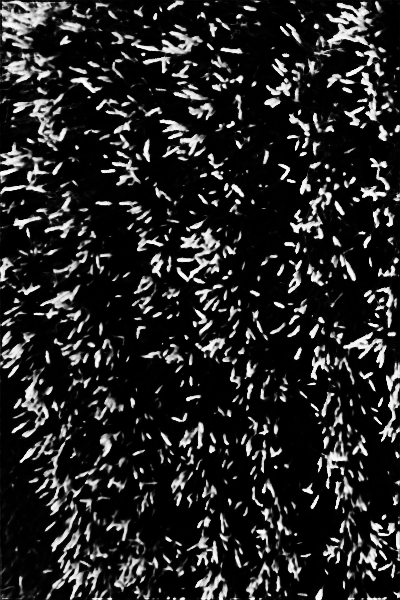}
        \includegraphics[width=0.32\textwidth]{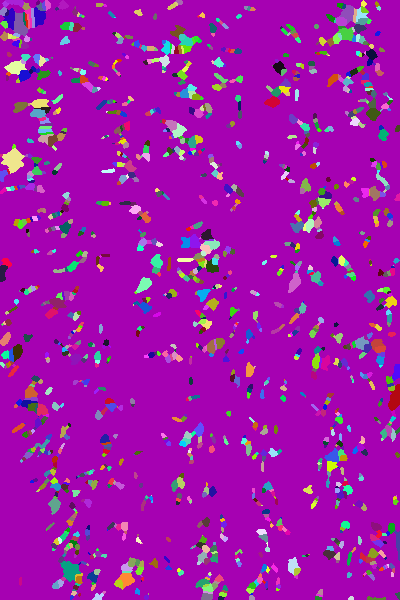}
    \end{subfigure}%
    \caption{Example image and intermediate outputs from the wheat dataset. Left to right: original image, softmax activations, watershed output.}
    \label{fig:examples-wheat}
\end{figure}

The proposed method was also tested on a dataset of wheat field images~(Table~\ref{tab:results}). The wheat head dataset consisted of 81 images from a \emph{T. aestivum} breeding trial captured with a camera extended from an all terrain vehicle pointed downward at a height of approximately 6~m. The images were captured in RAW format with a 18 MP camera with an ultra-wide angle rectilinear lens. Wheat head centers were annotated by domain experts using a custom tool and the total number of boxes was used for the annotated count value. The images were resized to 400 by 600 pixels for expediency because of the extremely high resolution, although this resizing is not required and the original resolution can be used as was the case for the sorghum datasets. Compared to the sorghum dataset, the wheat images included substantially more visual variation among images, including different head densities, different levels of maturity, and both awned and unawned heads. Figure~\ref{fig:examples-wheat} shows an example from the wheat dataset. Despite this variation, AutoCount was applied to the wheat dataset with no modifications. Although the detection of heads is positive, the arrangement of the heads in dense clumps proved difficult to segment accurately using the watershed method, resulting in a lower $\text{R}^2$ value than in the sorghum experiments. This issue is discussed further in Section~\ref{sec:discussion}.

\section{Discussion}
\label{sec:discussion}

\begin{figure}[h]
    \centering
    \includegraphics[width=\textwidth]{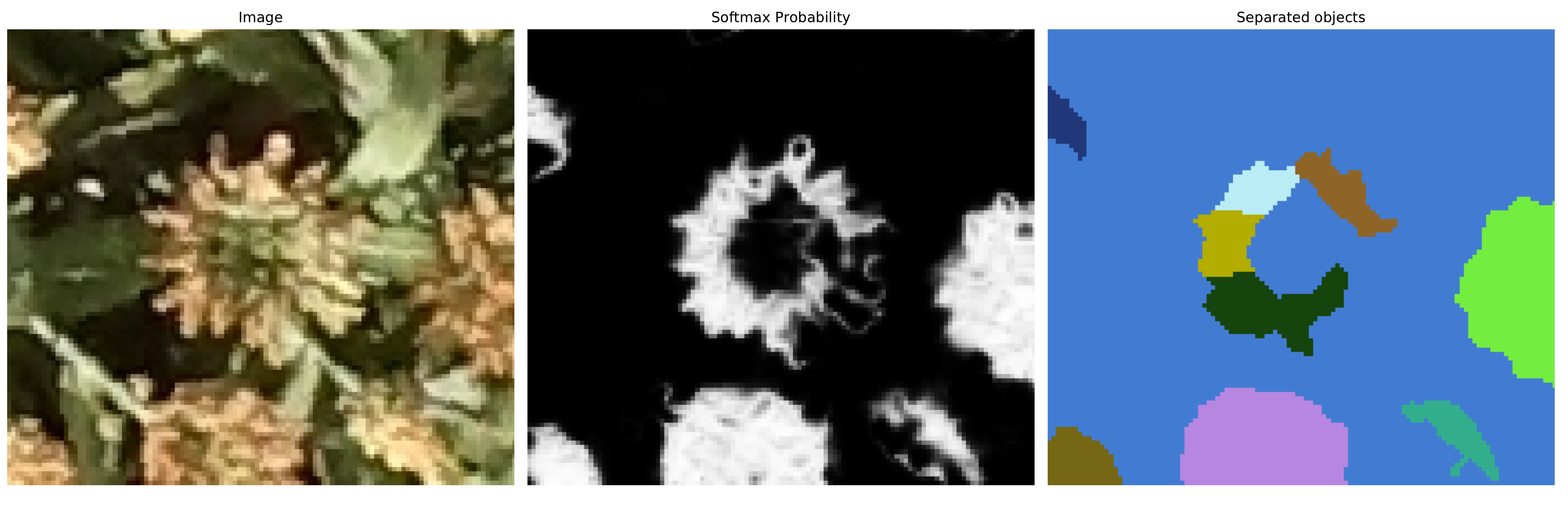}
    \caption{A failure case from the second sorghum dataset. Ring-shaped organs are over-segmented by the watershed method.}
    \label{fig:ring-fail}
\end{figure}

AutoCount is best suited to datasets where organs are represented as contiguous regions, making aerial imagery a good candidate. In contrast, high-resolution proximal imaging could create a scenario where the organ region is over-segmented, for example, by segmenting a wheat head into individual spikelets. The counting performance of the method is expected to suffer under these conditions. The method also is subject to the weaknesses of watershed segmentation. One such drawback is the potential over-segmentation of objects which are not compact. For example, the performance of the method suffered significantly in the second dataset of sorghum due to a small subset of images containing a high density of ring-shaped panicles. An example of this mode of failure is shown in Figure~\ref{fig:ring-fail}. Another example of a failure case for the watershed method can be seen in the dense clumps of wheat heads shown in Figure~\ref{fig:examples-wheat}. Many of these regions do not admit a shape which can be accurately segmented by the watershed method. Without the ability to accurately separate these large regions, the optimization of segmentation parameters ultimately settles on an over-segmentation of the wheat heads as can be seen in Figure~\ref{fig:distributions}c.

Although we focus on the organ count in the present work as it is the most agronomically relevant trait, the method also provides an instance segmentation of organs by way of the final watershed segmentation following optimization. This mask could potentially be used for other applications such as measuring organ size. Future work could potentially address replacing the watershed method with a more robust segmentation technique, such as one that is able to act on the activation map directly as opposed to requiring a thresholded input. We expect that a more robust segmentation technique would help boost performance for wheat head images and for other challenging object instances which tend to become over-segmented with our current approach, and could be dropped into the overall system without the need to change the segmentation loss function.

\section{Conclusion}

Here we have introduced the first unsupervised object counting algorithm specifically tailored for counting plant organs in field environments, called AutoCount. Although it does not rely on labelled data, the counting performance of the method is comparable to that of previously reported supervised methods on two datasets of sorghum. Additional positive results are presented on a new dataset of wheat heads. Some limitations of the method and avenues for future work are identified. The results presented here represent the first fully unsupervised baselines for performing plant organ counting of arbitrary plant organs in diverse field imagery.

\subsubsection*{Acknowledgments}
This research was funded by a Canada First Research Excellence Fund grant from the Natural Sciences and Engineering Research Council of Canada.

\bibliographystyle{splncs04}
\bibliography{paper}

\end{document}